\documentclass{article}

     \usepackage[nonatbib]{ml4eng_2020}

\usepackage[utf8]{inputenc} 
\usepackage[T1]{fontenc}    
\usepackage{hyperref}       
\usepackage{url}            
\usepackage{booktabs}       
\usepackage{amsfonts}       
\usepackage{nicefrac}       
\usepackage{microtype}      
\usepackage{graphicx}
\usepackage{amssymb}
\usepackage{amsmath}
\usepackage{mathtools}

\title{A General Framework Combining Generative Adversarial Networks and Mixture Density Networks for Inverse Modeling in Microstructural Materials Design}

\author{
  Zijiang Yang, Dipendra Jha, Arindam Paul, Wei-keng Liao, Alok Choudhary, Ankit Agrawal\\
  Department of Electrical and Computer Engineering\\
  Northwestern University\\
  \texttt{\{zyz293, dkj755, apx748, wkliao, choudhar, ankitag\}@ece.northwestern.edu} \\
}

\begin{document}

\maketitle

\begin{abstract}
Microstructural materials design is one of the most important applications of inverse modeling in materials science. Generally speaking, there are two broad modeling paradigms in scientific applications: forward and inverse. While the forward modeling estimates the observations based on known parameters, the inverse modeling attempts to infer the parameters given the observations. Inverse problems are usually more critical as well as difficult in scientific applications as they seek to explore the parameters that cannot be directly observed. Inverse problems are used extensively in various scientific fields, such as geophysics, healthcare and materials science. However, it is challenging to solve inverse problems, because they usually need to learn a one-to-many non-linear mapping, and also require significant computing time, especially for high-dimensional parameter space. Further, inverse problems become even more difficult to solve when the dimension of input (i.e. observation) is much lower than that of output (i.e. parameters). In this work, we propose a framework consisting of generative adversarial networks and mixture density networks for inverse modeling, and it is evaluated on a materials science dataset for microstructural materials design. Compared with baseline methods, the results demonstrate that the proposed framework can overcome the above-mentioned challenges and produce multiple promising solutions in an efficient manner.
\end{abstract}

\section{Introduction}
Microstructural materials design is one of the inverse modeling application in materials science, and it has revolutionarily changed the way advanced materials are developed \cite{thornton2009computational, kuehmann2009computational, agrawal2016perspective, olson1997computational}. However, there are still many challenges for inverse modeling, including microstructural materials design: 1) Inverse modeling usually requires learning a one-to-many non-linear mapping. Because it is possible that different input combinations from many causal factors might cause same output, this lack of uniqueness makes it difficult to train inverse models. 2) Inverse models usually need to learn a mapping from low-dimension inputs to high-dimension outputs, which means important missing information needs to be recovered from less informational inputs to produce high informational outputs. Thus, if the inverse model directly learns the mapping from inputs to outputs, the outputs might have limited diversity and only cover a small portion of real data distribution, especially when the difference of dimensionality between inputs and outputs is significant. 3) Traditional approaches for inverse modeling usually involves iterative learning process, such as optimization, so that optimal or near-optimal solution can gradually be achieved by minimizing the error between predicted solution and target. However, due to the fact that the input of a inverse model usually has much less dimension than the outputs, inverse modeling requires significant computing time.   

To overcome the above challenges, we propose a framework that combines generative adversarial networks (GAN) \cite{goodfellow2014generative} and mixture density networks (MDN) \cite{bishop1994mixture} for inverse modeling. Compared with three baseline methods, the results show that the proposed framework can not only generate solutions with property closer to the target property, but also produce more candidate solutions in an efficient manner
\footnote[1]{Code and data are available at: \url{https://github.com/zyz293/GAN-MDN}}.
 
\section{Methodology}
\subsection{Microstructural materials design}
Microstructural materials design is one of the inverse problems in the research of materials science. Microstructural materials design is the process to design materials microstructure to achieve a required property of the resulting material. Microstructural materials design has revolutionarily changed the way to design advanced materials and discover the new materials \cite{thornton2009computational}. In this work, we focus on the design of microstructure image with desired optical absorption property. Optical absorption property is defined as the ability of a materials to convert abosrted light into another energe form such as heat. Materials with high optical absorption property can be used in solar cell design.

\subsection{Proposed method}
The flowchart of the proposed method is shown in Figure \ref{flow} (a). The proposed method consists of GAN and MDN (see Appendix about details of GAN and MDN models) where GAN is used to obtain the low-dimensional design representations (i.e. latent variable vector) of the microstructure images and then MDN models the mapping between latent variable vector and design objective (i.e. material's optical absorption property). More specific, a GAN \cite{yang2018microstructural,li2018deep} is first trained so that the high-dimensional (i.e. high-resolution) image $x$ can be represented by low-dimensional latent variable vector $z$. Thus, we can utilize MDN (as shown in Figure \ref{flow} (b)), a neural network attempting to learn one-to-many non-linear mapping (i.e. solve challenge 1 mentioned in the introduction section), to model the mapping from image property $y$ to latent variable vector $z$ instead of directly mapping from image property $y$ to image $x$. Because latent variable vector $z$ has similar dimensionality as the image property $y$, it is easier and more stable to train the MDN by using latent variable vector $z$ as an immediate representations of image $x$ (i.e. solve challenge 2). Also, it is expected to increase the diversity of the outputs of the inverse model to cover wider range of real data distribution. After the proposed framework is well-trained, given a desired image property $y$, the MDN can produce various sets of latent variable vector $z$, which can be further used by GAN to generate corresponding images $x$ to solve the inverse problem. Because the proposed framework is based on deep learning, it only requires one forward pass to produce various predictions, which means it only needs few seconds to produce possible solutions using modern computation resources (i.e. solve challenge 3).

\begin{figure}[htbp]
\centerline{\includegraphics[scale=0.57]{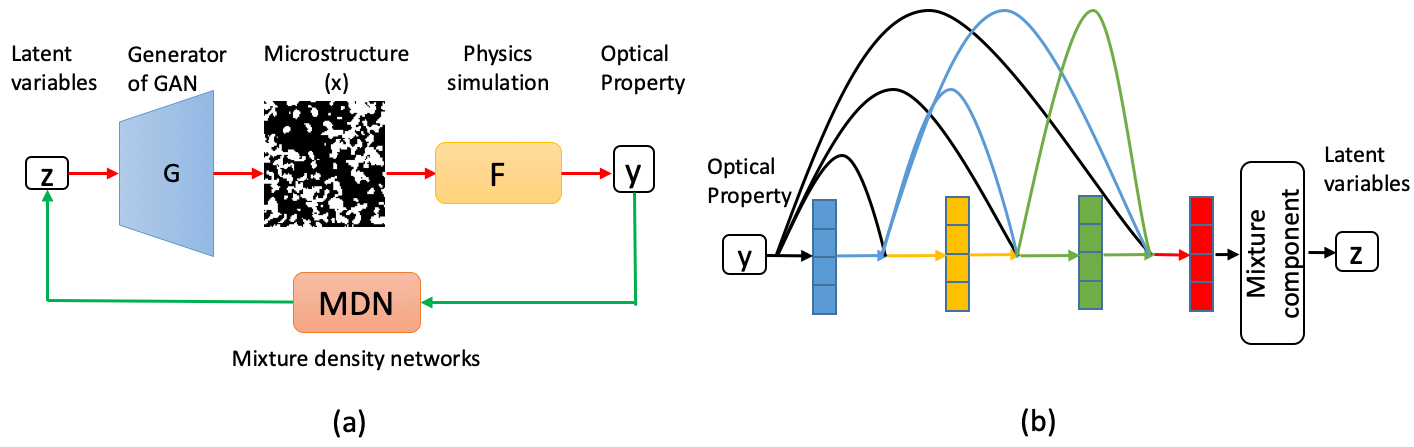}}
\caption{(a) The flowchart of the proposed method. The red path shows the flow of data generation, and the green path represents the training of the proposed MDN. (b) The architecture of the proposed MDN. MDN is constructed by four densely connected fully connected layers and a mixture component that models a mixture of Gaussian distributions.}
\label{flow}
\end{figure}

\section{Experiments}
\subsection{Datasets and error metric}
In \cite{yang2018microstructural,li2018deep}, the GAN is first trained on synthetic microstructure images, which are created using Gaussian Random Field (GRF) method. The parameters in GRF (i.e. mean, standard deviation, and volume fraction) are carefully controlled to produce microstructures that cover the vast space of compositional and dispersive patterns, which corresponds to different processing conditions of the same material system. Then, the GAN and physics simulation in \cite{yang2018microstructural,li2018deep, yu2017characterization} are used to generate two datasets used in this work, and the data generation flow of which is shown in the red path of Figure \ref{flow} (a). The size of microstructure image $x$ and the latent variable vector $z$ of one dataset are $96\times{96}$ and $3\times{3}$ (referred as Data-I), and the other is $64\times{64}$ and $2\times{2}$ (referred as Data-II), respectively. More specifically, the latent variable vector $z$ is randomly generated and passed through generator to generate corresponding microstructure $x$. Then the optical absorption property $y$ of the generated microstructure is simulated using physics simulation (i.e. the rigorous coupled wave analysis \cite{yu2017characterization}). Then the two datasets could be used to train the proposed MDN to learn the mapping between latent variable vector $z$ and optical absorption property $y$ as shown in the green path in Figure \ref{flow} (a), respectively.

Residual error percentage (REP) is used to evaluate the performance of models, which is defined as equation \ref{e1},
\begin{equation}
\label{e1}
REP = \frac{\mid \hat{y} - y \mid}{y} \times 100 \%
\end{equation}
where $\hat{y}$ and $y$ are the optical absorption property of generated microstructure and target optical absorption property, respectively. 

\subsection{Baselines}
\textbf{Optimization based inverse modeling:} The inverse modeling method based on optimization in \cite{yang2018microstructural,li2018deep} is considered as a baseline method in this work. More specifically, for each target optical absorption property, 250 sample pairs $(z, y)$ are sampled in the design representations (i.e. latent variable vector of GAN) space to create the response surface between latent variable vector and materials optical absorption property. Then, metamodel-based Bayesian optimization is conducted to iteratively explore the next potentially optimal design point. 400 iterations of optimization are conducted after initial sampling of 250 points to ensure the convergence of the optimization process.  

\textbf{MDN based deep learning inverse modeling:} In order to illustrate that it is easier and more stable to learn the mapping from materials optical absorption property $y$ to latent variable vector $z$ of GAN instead of directly mapping from materials optical absorption property $y$ to microstructural images $x$, we use a deep learning inverse modeling solely based on MDN as another baseline. More specifically, MDN takes materials optical absorption property $y$ as input and directly produces microstructural images $x$. The MDN in this baseline is the same as the MDN in proposed framework, except the number of neurons in the mixture component is different because the dimension of the output is different (i.e. each pixel in microstrcuture image $x$ can be considered as one dimension). Other hyperparameter settings and training strategy are the same as the proposed framework.

\textbf{PCA and MDN based inverse modeling (referred as PCA-MDN method):} In order to illustrate the advantage of using GAN to obtain the low-dimensional design representations (i.e. latent variable vector) as compared to traditional dimensionality reduction methods, PCA is used to replace GAN and combined with MDN to produce microstructure images $x$ given a desired materials optical absorption property $y$. More specifically, MDN takes materials optical absorption property $y$ as input and generates reduced set of principal components, which is used by PCA to inversely transform to corresponding microstructure images $x$. The dimension of reduced set of principal components is the same as the dimension of latent variable vector in proposed framework, and MDN is also exactly the same as the MDN in proposed framework.

\subsection{Results and discussion}
We select five target optical absorption properties (i.e. 0.55, 0.60, 0.65, 0.70 and 0.75) to cover the range of possible optical absorption properties. For each target optical absorption property, we use the proposed MDN to sample 30 latent variable vectors $z$ where we randomly select a distribution based on its mixture coefficient to sample the latent variable vectors. Each latent variable vector $z$ is then passed through the proposed GAN to generate microstructure images. Finally, their corresponding optical absorption property can be simulated by physics simulation \cite{yang2018microstructural,li2018deep, yu2017characterization} and compared with target optical absorption property. The same evaluation strategy is used for the baseline methods (i.e. MDN based deep learning inverse modeling, and PCA-MDN method).

\begin{figure}[htbp]
\centerline{\includegraphics[scale=0.57]{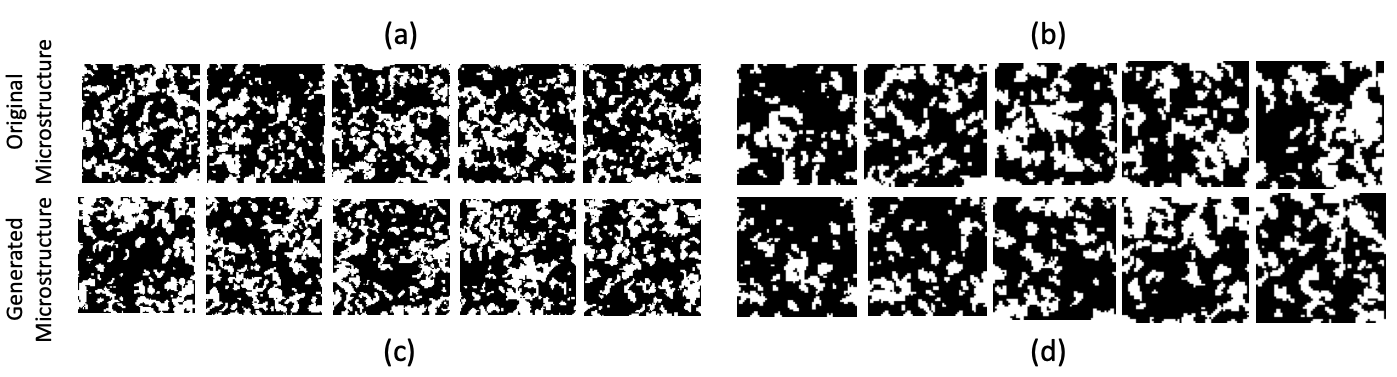}}
\caption{Examples of original microstructures and microstructures produced by the proposed MDN for Data-I and Data-II. Row (a) and (b) are microstructures randomly selected in Data-I and Data-II. Row (c) and (d) are microstructures produced by the proposed MDN with minimun REP to the target optical absorption properties. The target optical absorption properties are 0.55, 0.60, 0.65, 0.70 and 0.75 from left to right.}
\label{data}
\end{figure}

\textbf{Results on Data-I:} Table \ref{table1} (left) shows the performance of the proposed framework on Data-I. We can observe that average REPs of the proposed method are the lowest for most target values. Moreover, min REPs are much less than 1\%, which is much lower than baselines. This indicates that the proposed method can generate microstructure with optical absorption property very close to target property. Figure \ref{data} (a) and (c) plots the examples of original microstructures and microstructures produced by the proposed MDN that have the min REP compared with each target optical absorption property for Data-I, respectively. It shows that the proposed MDN is capable of producing latent variable vectors $z$ that generate visually similar microstructures as the original microstructures in the dataset. In addition, it only takes around 10 seconds to produce designed microstructural images.

\begin{table}[htbp]
\scriptsize
\caption{Performance of the proposed method, MDN based deep learning inverse modeling baseline method and optimization based inverse modeling baseline method on Data-I (left) and Data-II (right)}
\begin{tabular}{ll}
\begin{minipage}{.5\textwidth}
\centering
\begin{tabular}{|c|p{0.7cm}|p{0.7cm}|p{1cm}|p{1cm}|}
\hline
\textbf{Value}  & \textbf{Min REP} & \textbf{Average REP} & \textbf{Standard deviation of REP} & \textbf{Running Time} \\\hline
\multicolumn{5}{|c|}{\textbf{The proposed method}} \\\hline
\textbf{0.55} & 0.65\% & 15.68\% & 8.40\% & 9.75s \\\hline
\textbf{0.60} & 0.18\% & 9.15\% & 5.97\% & 9.50s  \\\hline
\textbf{0.65} & 0.22\% & 5.80\% & 3.93\% & 9.67s \\\hline
\textbf{0.70} & 0.13\% & 5.29\% & 3.86\% & 9.62s \\\hline
\textbf{0.75} & 0.20\% & 7.83\% & 3.91\% & 9.50s \\\hline
\multicolumn{5}{|c|}{\textbf{Baseline: PCA-MDN method}} \\\hline
\textbf{0.55} & 5.05\% & 17.67\% & 7.84\% & 7.22s \\\hline
\textbf{0.60} & 0.50\% & 10.89\% & 6.48\% & 7.30s \\\hline
\textbf{0.65} & 0.17\% & 5.92\% & 4.00\% & 7.20s \\\hline
\textbf{0.70} & 0.40\% & 8.81\% & 5.27\% & 7.20s \\\hline
\textbf{0.75} & 2.95\% & 18.34\% & 5.54\% & 7.36s \\\hline
\multicolumn{5}{|c|}{\textbf{Baseline: MDN based deep learning inverse modeling}} \\\hline
\textbf{0.55} & 0.84\% & 9.07\% & 3.14\% & 175.27s \\\hline
\textbf{0.60} & 4.70\% & 14.40\% & 4.08\% & 187.86s \\\hline
\textbf{0.65} & 9.35\% & 20.04\% & 4.06\% & 177.60s \\\hline
\textbf{0.70} & 12.29\% & 25.18\% & 4.21\% & 147.23s \\\hline
\textbf{0.75} & 17.73\% & 26.81\% & 3.55\% & 178.70s \\\hline
\multicolumn{5}{|c|}{\textbf{Baseline: Optimization based inverse modeling}} \\\hline
\textbf{0.55} & - & - & - & 4.4h \\\hline
\textbf{0.60} & 1.08\% & - & - & 3.6h \\\hline
\textbf{0.65} & 3.38\% & - & - & 5.8h \\\hline
\textbf{0.70} & - & - & - & 10.6h \\\hline
\textbf{0.75} & - & - & - & 8.9h \\\hline
\end{tabular}
\end{minipage} 

\begin{minipage}{.5\textwidth}
\centering
\begin{tabular}{|c|p{0.7cm}|p{0.7cm}|p{1cm}|p{1cm}|}
\hline
\textbf{Value} & \textbf{Min REP} & \textbf{Average REP} & \textbf{Standard deviation of REP} & \textbf{Running Time} \\\hline
\multicolumn{5}{|c|}{\textbf{The proposed method}} \\\hline
\textbf{0.55} & 1.25\% & 16.19\% & 8.96\% & 9.67s \\\hline
\textbf{0.60} & 0.70\% & 10.99\% & 7.93\% & 9.74s \\\hline
\textbf{0.65} & 0.18\% & 7.65\% & 5.64\% & 9.57s \\\hline
\textbf{0.70} & 0.10\% & 5.00\% & 4.61\% & 9.68s \\\hline
\textbf{0.75} & 0.43\% & 6.18\% & 3.51\% & 9.60s \\\hline
\multicolumn{5}{|c|}{\textbf{Baseline: PCA-MDN method}} \\\hline
\textbf{0.55} & 4.96\% & 11.74\% & 3.05\% & 7.24s \\\hline
\textbf{0.60} & 0.07\% & 2.69\% & 2.18\% & 7.26s \\\hline
\textbf{0.65} & 3.71\% & 8.79\% & 2.59\% & 7.40s \\\hline
\textbf{0.70} & 0.10\% & 3.41\% & 2.44\% & 7.15s \\\hline
\textbf{0.75} & 3.17\% & 6.27\% & 1.52\% & 7.26s \\\hline
\multicolumn{5}{|c|}{\textbf{Baseline: MDN based deep learning inverse modeling}} \\\hline
\textbf{0.55} & 2.85\% & 12.78\% & 3.89\% & 23.21s \\\hline
\textbf{0.60} & 7.87\% & 14.95\% & 3.56\% & 24.05s \\\hline
\textbf{0.65} & 11.00\% & 17.33\% & 2.63\% & 24.14s \\\hline
\textbf{0.70} & 3.03\% & 15.62\% & 4.09\% & 23.90s \\\hline
\textbf{0.75} & 8.44\% & 12.73\% & 3.20\% & 23.34s \\\hline
\multicolumn{5}{|c|}{\textbf{Baseline: Optimization based inverse modeling}} \\\hline
\textbf{0.55} & 15.51\% & - & - & 5.8h \\\hline
\textbf{0.60} & - & - & - & 12.1h \\\hline
\textbf{0.65} & 1.21\% & - & - & 4.2h \\\hline
\textbf{0.70} & - & - & - & 18.8h \\\hline
\textbf{0.75} & - & - & - & 3.2h \\\hline
\end{tabular}
\end{minipage}
\end{tabular}
\label{table1}
\end{table}

The results of PCA-MDN method is also shown in Table \ref{table1} (left). We can observe that the proposed method has significantly better min REP and average REP than PCA-MDN method for low and high target optical absorption property values (i.e. 0.55, 0.70 and 0.75). This might indicate that this baseline method is incapable to capture all the significant information, so it fails to generate microstructures with properties exactly matching the target properties since PCA can lose more important information when obtaining low-dimensional design representations compared to GAN. Similarly to the proposed method, it only takes a few seconds to produce microstructures.

Table \ref{table1} (left) also shows the performance of MDN based deep learning inverse modeling baseline method. The results show that both min REPs and average REPs are much higher than that of the proposed framework for most target values, and REP increases with the target optical absorption property increasing. This is because this baseline method mainly produces microstructure images with low optical absorption property. In other words, this baseline method focuses more on the low optical absorption property range of real data distribution. Thus, it fails to generate possible microstructure images when target optical absorption property is high. As discussed in the introduction section, the significant difference of dimensionality between optical absorption property $y$ and microstructure images $x$ makes the training of inverse model even more difficult. Thus, the diversity of the generated microstructure images is limited and only a small portion of real data distribution is covered by directly modeling the relationship between optical absorption property and microstructure images. In contrast, by using the latent variable vector $z$ as the immediate representations of microstructure images $x$, the proposed framework provides diverse microstructure images along the entire range of optical absorption properties. Although this baseline is also based on deep learning, it takes more time to produce microstructures compared to the proposed method since it directly maps optical absorption property $y$ to microstructure images $x$.

\begin{figure*}[htbp]
\centerline{\includegraphics[scale=0.57]{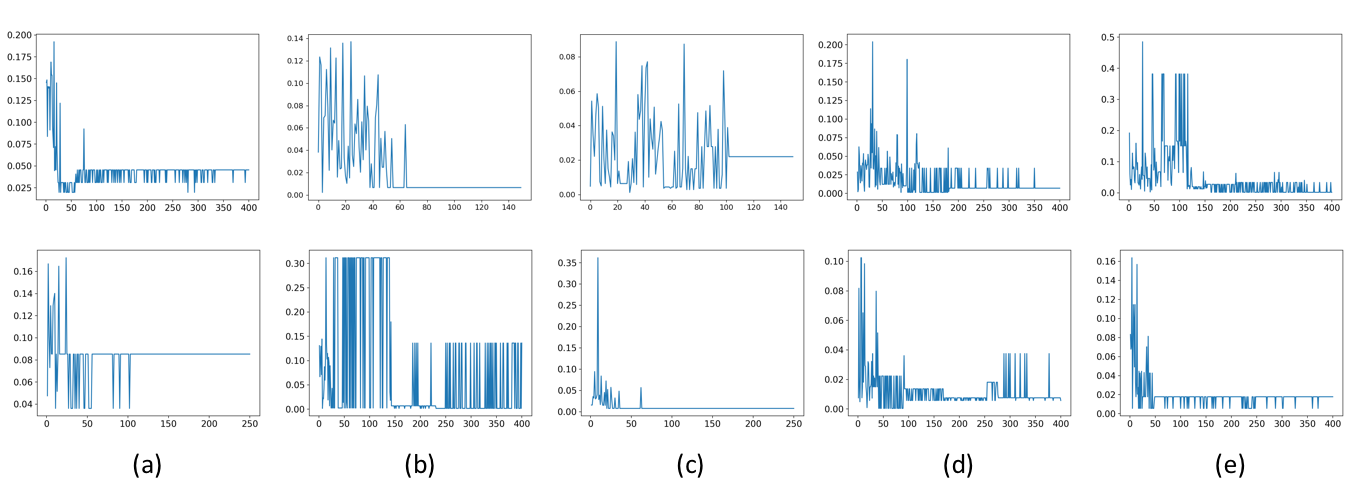}}
\caption{The microstructure optimization history for each target optical absorption property for Data-I (i.e. first row) and Data-II (i.e. second row). The x axis shows the iteration number, and the y axis shows the absolute REP between target optical absorption property and optical absorption property of sampled microstructure. The target optical absorption properties of each plots are 0.55 (a), 0.60 (b), 0.65 (c), 0.70 (d) and 0.75 (e).}
\label{optim}
\end{figure*}

The performance of optimization based baseline method on Data-I is also listed in Table \ref{table1} (left), and the first row in Figure \ref{optim} shows the microstructure optimization history for each target optical absorption property. Since optimization method can only produce one candidate microstructure, the average and standard deviation of REP is not applicable. For target properties 0.6 and 0.65, this baseline method reaches convergence around 65 and 105 epochs, respectively. However, this baseline method cannot converge when target properties are 0.55, 0.7 and 0.75. We can observe significant advantages of the proposed framework because optimization based method could not get comparable performance as the proposed method or even could not converge. The results indicate that optimization based inverse modeling cannot successfully capture the relationship between latent variable vector $z$ and optical absorption property $y$ and is incapable to generate microstructure with desired property for all values. In addition, optimization based inverse modeling can only produce limited number of candidate microstructures due to the nature of optimization method, while the proposed framework can sample as many candidate microstructures as user wants. More importantly, it takes hours for optimization based inverse modeling to optimize the microstructure for desired optical absorption property, while it only needs one forward pass for the proposed framework to produce microstructures, which only takes around 10 seconds using a Titan X GPU. 

\textbf{Results on Data-II:} Table \ref{table1} (right) shows the performance of the proposed framework on Data-II. The min REP and average REP for each target optical absorption property is extremely small, and the performance is comparable with that on Data-I. In addition, we can observe in Figure \ref{data} (c) and (d) that the proposed framework can generate visually similar microstructures as microstructures in the dataset. 

The results of PCA-MDN method is also shown in Table \ref{table1} (right). The results of PCA-MDN method are comparable with the proposed method, and it achieves better average REP in some cases. This might because the microstructure image in Data-II is smaller than that in Data-I so that PCA is capable to capture enough information. However, the min REP of PCA-MDN method is significant worse for some target properties, which might indicate it is not stable to use PCA to obtain low-dimensional design representations compared to GAN. In addition, similar as Data-I, it only takes a few seconds for PCA-MDN method to produce microstructures.

Table \ref{table1} (right) presents the performance of MDN based deep learning inverse modeling method, and the performance is much worse than that of the proposed framework. Compared to its performance on Data-I, it also focuses more on the low optical absorption property range of real data distribution, but it covers a wider property range since it performs better on high property values on Data-II. This is because the microstructure image in Data-II is smaller than that in Data-I so that it is easier for MDN to directly map from optical absorption property $y$ to microstructure image $x$. This observation also supports our conclusion that it is easier and more stable to train the MDN by using latent variable vector $z$ as an immediate representations of image $x$ since the performance of the proposed framework on two datasets are similar. Since microstructure images in Data-II is smaller, it takes less time to produce microstructures compared to running time in Data-I.

The second row in Figure \ref{optim} shows the optimization history of the optimization based baseline method on Data-II. We can observe that when the target properties are 0.55 and 0.65, the baseline method reaches convergence around 110 and 70 epochs, respectively. However, it fails to converge when target properties are 0.6, 0.7 and 0.75. The performance of optimization based inverse modeling method is presented in Table \ref{table1} (right). It shows that the performance of the proposed framework is much better than that of optimization based method. More importantly, it still takes hours to produce the solutions even though the dimension of latent variable vector to be optimized for Data-II is smaller than that for Data-I, and it is much slower than the proposed framework.

\section{Conclusion}
In this work, we propose a framework that combines generative adversarial networks and mixture density networks for inverse modeling in materials microstructural design. Compared with three benchmark methods, the proposed framework provides a more stable way to solve inverse problems. The results show that the proposed framework can not only generate microstructures that have property closer to the target property, but also it can produce more candidate microstructures in an efficient manner.

\section*{Acknowledgment}
The Rigorous Couple Wave Analysis simulation is supported by Prof. Cheng Sun’s lab at Northwestern University. This work is supported in part by the following grants: NIST award 70NANB19H005; DOE awards DE-SC0014330, DE-SC0019358.

\bibliographystyle{bmc-mathphys}
\bibliography{reference.bib}

\section*{Appendix}
\subsection*{Generative adversarial network}
Generative adversarial networks \cite{goodfellow2014generative} is a type of deep learning system originated from game theory. GAN consists of two components: generator and discriminator. Specifically, generator $G(z)$ produces samples $x_G$ from latent variable vector $z$ to approximate samples $x_{data}$ from real dataset, while discriminator $D(x)$ distinguishes the generated samples $x_G$ from real samples $x_{data}$. Essentially, GAN is defined as a minimax game, which can be formulated as following equation,
\begin{equation}
\min_{G} \max_{D} V(D,G) =
\mathbb{E}_{x\sim{p_{data}(x)}}[\log{D(\textbf{x})}]
+ \mathbb{E}_{\textbf{z}\sim{p_z(z)}}[\log(1-D(G(\textbf{z})))]
\label{minmax_GAN}
\end{equation}
where ${p_z(z)}$ is the prior distribution of the latent variable vector $z$, and ${p_{data}(x)}$ is the distribution of the real data $x_{data}$. This minimax game would eventually lead to a convergence where generator can generate data similar to real data that cannot be distinguished by discriminator. 

The GAN in this work is a fully convolutional neural network where both generator and discriminator have five layers. Specifically, each generator layer is a deconvolutional layer attached with batch normalization (BN) operation and rectified linear unit (ReLU) activation, except the last layer which uses a tanh activation function to produce the bounded pixel values for generated images. The number of filters in the five deconvolutional layers are 128, 64, 32, 16 and 1, respectively. Each discriminator layer consists of convolutional layer, BN operation and leaky rectified linear unit activation, except the last layer which uses a sigmoid activation function to predict whether the image is fake or real. The number of filters in the five convolutional layers are 16, 32, 64, 128 and 1, respectively. For both convolutional and deconvolutional layers, the filter size is $4\times{4}$ with stride 2, except the last convolutional layer in discriminator where the filter size is the same as the size of its input feature maps to produce probabilities. In order to avoid model collapse and impose morphology constraints of the generated images, model collapse loss and style transfer loss are added in addition to adversarial loss. (i.e. see \cite{yang2018microstructural, li2018deep} for details about customized loss function)

\subsection*{Mixture density network}
Mixture density networks \cite{bishop1994mixture} is a type of neural network attempting to address inverse problem. Instead of predicting a single value, the goal of MDN is to predict an entire probability distribution for the output (i.e. latent variable vector $z$) based on input (i.e. optical absorption property value $y$). Specifically, MDN is usually constructed by a neural network to parameterize a mixture model consisting of some predefined distributions. Generally, Gaussian distribution is used, and the output is modeled as a conditional probability $P(z|y)$ calculated by a weighted sum of $K$ gaussian distributions $\phi$ with different means $\mu$ and standard deviations $\sigma$, which can be defined as following.
\begin{equation}
P(z|y) = \sum_{k=1}^{K} \pi_{k}(y)\phi(z|\mu_k(y),\sigma_k(y)), \sum_{k=1}^{K} \pi_{k}(y)=1
\label{mdn}
\end{equation}
where $y$ and $z$ are inputs and outputs, respectively. $\pi_k$, $\mu_k$ and $\sigma_k$ are the mixing coefficient, mean and standard deviation of the $k^{th}$ gaussian distribution, respectively. The network is updated by minimizing the logarithm of the likelihood of the distribution versus the training data, 
\begin{equation}
\min L(w) = \frac{-1}{N}\sum_{n=1}^{N}\log(
\sum_{k}\pi_{k}(y_n, w)\phi(z_n|\mu_k(y_n,w),\sigma_k(y_n,w)))
\label{loss_mdn}
\end{equation}
where $N$ is the batch size, $w$ is the weights of the MDN, $y_n$ is the $n^{th}$ instance in a batch and $z_n$ is the corresponding label.

MDN in this work is constructed by four densely connected fully connected layers and a mixture component that models a mixture of Gaussian distributions. Each densely connected fully connected layer has 16 neurons followed by BN operation and ReLU activation, and each layer is connected with subsequent layers. In other words, each layer obtains additional inputs from all preceding layers (including input layer) to calculate its outputs. The mixture component contains a mixture of 40 multivariate Gaussian distributions, which are parameterized by a fully connected layer. Assuming $M$ denotes the dimension of the output (i.e. the dimension of latent variable vector of GAN), each multivariate Gaussian distribution needs one neuron for its mixing coefficient, two neurons for mean and standard deviation of each dimension of latent variable vector. In particular, linear activation function is used for the neurons computing mixing coefficients and means, while exponential linear unit (ELU) \cite{clevert2015fast} is used for the neurons calculating standard deviations. Thus, the number of neurons in the mixture component is computed as following,
\begin{equation}
N = K\times{(1+2\times{M})}
\label{num_neuron}
\end{equation}
where $N$ is the number of neurons of the mixture component, $K$ is the number of multivariate Gaussian distributions used in MDN ($K=40$ in the proposed model). During prediction time, we randomly select a distribution based on its mixture coefficient to sample the latent variable vectors. For MDN training, we use equation \ref{loss_mdn} as the loss function. Adam optimizer \cite{kingma2014adam} with batch size of 128 and learning rate of 0.001 is used. Early stopping with a patience of 50 is applied so that the training process is terminated when the loss function on validation set does not improve for 50 epochs. 

\end{document}